\documentclass{article}

\makeatletter

\usepackage[verbose=true,letterpaper]{geometry}
\AtBeginDocument{
  \newgeometry{
    textheight=9in,
    textwidth=6.5in,
    top=1in,
    headheight=14pt,
    headsep=25pt,
    footskip=30pt
  }
}

\usepackage{fancyhdr}
\fancyheadoffset{0pt}
\def\keywordname{{\bfseries \emph Keywords}}%
\def\keywords#1{\par\addvspace\medskipamount{\rightskip=0pt plus1cm
\def\and{\ifhmode\unskip\nobreak\fi\ $\cdot$
}\noindent\keywordname\enspace\ignorespaces#1\par}}

\renewcommand{\normalsize}{%
  \@setfontsize\normalsize\@xpt\@xipt
  \abovedisplayskip      7\p@ \@plus 2\p@ \@minus 5\p@
  \abovedisplayshortskip \z@ \@plus 3\p@
  \belowdisplayskip      \abovedisplayskip
  \belowdisplayshortskip 4\p@ \@plus 3\p@ \@minus 3\p@
}
\normalsize
\renewcommand{\small}{%
  \@setfontsize\small\@ixpt\@xpt
  \abovedisplayskip      6\p@ \@plus 1.5\p@ \@minus 4\p@
  \abovedisplayshortskip \z@  \@plus 2\p@
  \belowdisplayskip      \abovedisplayskip
  \belowdisplayshortskip 3\p@ \@plus 2\p@   \@minus 2\p@
}
\renewcommand{\footnotesize}{\@setfontsize\footnotesize\@ixpt\@xpt}
\renewcommand{\scriptsize}{\@setfontsize\scriptsize\@viipt\@viiipt}
\renewcommand{\tiny}{\@setfontsize\tiny\@vipt\@viipt}
\renewcommand{\large}{\@setfontsize\large\@xiipt{14}}
\renewcommand{\Large}{\@setfontsize\Large\@xivpt{16}}
\renewcommand{\LARGE}{\@setfontsize\LARGE\@xviipt{20}}
\renewcommand{\huge}{\@setfontsize\huge\@xxpt{23}}
\renewcommand{\Huge}{\@setfontsize\Huge\@xxvpt{28}}

\providecommand{\section}{}
\renewcommand{\section}{%
  \@startsection{section}{1}{\z@}%
                {-2.0ex \@plus -0.5ex \@minus -0.2ex}%
                { 1.5ex \@plus  0.3ex \@minus  0.2ex}%
                {\large\bf\raggedright}%
}
\providecommand{\subsection}{}
\renewcommand{\subsection}{%
  \@startsection{subsection}{2}{\z@}%
                {-1.8ex \@plus -0.5ex \@minus -0.2ex}%
                { 0.8ex \@plus  0.2ex}%
                {\normalsize\bf\raggedright}%
}
\providecommand{\subsubsection}{}
\renewcommand{\subsubsection}{%
  \@startsection{subsubsection}{3}{\z@}%
                {-1.5ex \@plus -0.5ex \@minus -0.2ex}%
                { 0.5ex \@plus  0.2ex}%
                {\normalsize\bf\raggedright}%
}
\providecommand{\paragraph}{}
\renewcommand{\paragraph}{%
  \@startsection{paragraph}{4}{\z@}%
                {1.5ex \@plus 0.5ex \@minus 0.2ex}%
                {-1em}%
                {\normalsize\bf}%
}
\providecommand{\subparagraph}{}
\renewcommand{\subparagraph}{%
  \@startsection{subparagraph}{5}{\z@}%
                {1.5ex \@plus 0.5ex \@minus 0.2ex}%
                {-1em}%
                {\normalsize\bf}%
}

\newlength{\@abovecaptionskip}
\newlength{\@belowcaptionskip}

\renewenvironment{table}
  {\setlength{\abovecaptionskip}{\@belowcaptionskip}%
   \setlength{\belowcaptionskip}{\@abovecaptionskip}%
   \@float{table}}
  {\end@float}

\renewcommand{\footnoterule}{\kern-3\p@ \hrule width 12pc \kern 2.6\p@}
\newcommand{\equalcontribnote}[1]{%
  \gdef\@equalcontribtext{#1}
}
\newcommand{\projectleadernote}[1]{%
  \gdef\@projectleadertext{#1}
}

\def\@equalcontribtext{}
\def\@projectleadertext{}

\def\@listi  {\leftmargin\leftmargini}
\def\@listii {\leftmargin\leftmarginii
              \labelwidth\leftmarginii
              \advance\labelwidth-\labelsep
              \topsep  2\p@ \@plus 1\p@    \@minus 0.5\p@
              \parsep  1\p@ \@plus 0.5\p@ \@minus 0.5\p@
              \itemsep \parsep}
\def\@listiii{\leftmargin\leftmarginiii
              \labelwidth\leftmarginiii
              \advance\labelwidth-\labelsep
              \topsep    1\p@ \@plus 0.5\p@ \@minus 0.5\p@
              \parsep    \z@
              \partopsep 0.5\p@ \@plus 0\p@ \@minus 0.5\p@
              \itemsep \topsep}
\def\@listiv {\leftmargin\leftmarginiv
              \labelwidth\leftmarginiv
              \advance\labelwidth-\labelsep}
\def\@listv  {\leftmargin\leftmarginv
              \labelwidth\leftmarginv
              \advance\labelwidth-\labelsep}
\def\@listvi {\leftmargin\leftmarginvi
              \labelwidth\leftmarginvi
              \advance\labelwidth-\labelsep}

\providecommand{\maketitle}{}
\renewcommand{\maketitle}{%
  \par
  \begingroup
    \renewcommand{\thefootnote}{\fnsymbol{footnote}}
    \renewcommand{\@makefnmark}{\hbox to \z@{$^{\@thefnmark}$\hss}}
    \long\def\@makefntext##1{%
      \parindent 1em\noindent
      \hbox to 1.8em{\hss $\m@th ^{\@thefnmark}$}##1
    }
    \thispagestyle{empty}
    \@maketitle
    \@thanks

    \ifx\@equalcontribtext\@empty
    \else
      \footnotetext[1]{\textit{\underline{\@equalcontribtext}}}
    \fi

    \ifx\@projectleadertext\@empty
    \else
      \footnotetext[2]{\textit{\underline{\@projectleadertext}}}
    \fi
  \endgroup
  \let\maketitle\relax
  \let\thanks\relax
}

\newcommand{\@toptitlebar}{
  \hrule height 2\p@
  \vskip 0.25in
  \vskip -\parskip%
}
\newcommand{\@bottomtitlebar}{
  \vskip 0.29in
  \vskip -\parskip
  \hrule height 2\p@
  \vskip 0.09in%
}

\providecommand{\@maketitle}{}
\renewcommand{\@maketitle}{%
  \vbox{%
    \hsize\textwidth
    \linewidth\hsize
    \vskip 0.1in
    \@toptitlebar
    \centering
    {\LARGE\sc \@title\par}
    \@bottomtitlebar
    \vskip 0.1in
    \def\And{%
      \end{tabular}\hfil\linebreak[0]\hfil%
      \begin{tabular}[t]{c}\bf\rule{\z@}{24\p@}\ignorespaces%
    }
    \def\AND{%
      \end{tabular}\hfil\linebreak[4]\hfil%
      \begin{tabular}[t]{c}\bf\rule{\z@}{24\p@}\ignorespaces%
    }
    \begin{tabular}[t]{c}\bf\rule{\z@}{24\p@}\@author\end{tabular}%
  \vskip 0.4in \@minus 0.1in \center{ }   \vskip 0.2in
  }
}

\newcommand{\ftype@noticebox}{8}
\newcommand{\@notice}{%
  \enlargethispage{2\baselineskip}%
  \@float{noticebox}[b]%
    \footnotesize\@noticestring%
  \end@float%
}

\renewenvironment{abstract}
{
  \centerline
  {\large \bfseries \scshape Abstract}
  \begin{quote}
}
{
  \end{quote}
}
\makeatother

\usepackage[utf8]{inputenc}
\usepackage[T1]{fontenc}
\usepackage{amsmath,amssymb,amsfonts}
\usepackage{url}
\usepackage[hidelinks]{hyperref}
\usepackage{microtype}
\usepackage{fancyhdr}
\usepackage{graphicx}
\usepackage{changepage}
\usepackage{cite}
\usepackage[labelfont=bf,labelsep=period,justification=raggedright,singlelinecheck=off]{caption}

\title{Knowledge distillation across teacher architectures for compact diabetic retinopathy grading}

\author{
  Mossa Manisa, Habiba Jahan Priya, A.K.M. Fakhrul Hossain \\
  Department of Computer Science and Engineering \\
  Shahjalal University of Science and Technology \\
  Sylhet, Bangladesh \\
  Corresponding author: \texttt{mossamanisa@gmail.com}
}

\begin{document}
\maketitle

\begin{abstract}
Diabetic retinopathy grading from retinal fundus photographs is important but difficult to deploy on resource-constrained hardware. We conducted a controlled study of offline response-based knowledge distillation for five-class diabetic retinopathy grading on APTOS 2019, comparing ResNet-50, MobileNetV3-Large, and Swin Transformer-Tiny (Swin-T) teachers distilled into a MobileNetV3-Small student. A $3 \times 4 \times 2$ grid of teacher, temperature, and hard-label weight was screened at one seed, while six configurations and the no-distillation baseline were evaluated across three seeds. The highest observed single-seed quadratic weighted kappa was 0.9093 (macro-F1 0.6851) for Swin-T at temperature 4 and hard-label weight 0.7; the corresponding three-seed mean was 0.9024, compared with 0.8804 for the baseline. The student used 1.52~M parameters and 5.94~MB, representing an 18.07$\times$ parameter reduction relative to Swin-T. Temperature and hard-label weight effects were teacher- and configuration-dependent, while small performance differences were sensitive to random seed. Knowledge distillation produced a compact student with competitive performance under the evaluated protocol, but independent external validation remains necessary.
\end{abstract}


\section{Introduction}

Diabetic retinopathy (DR) is one of the leading causes of preventable visual impairment worldwide, affecting an estimated one-third of all individuals with diabetes mellitus~\cite{yau2012global}. DR is clinically classified into five severity levels such as No DR, Mild, Moderate, Severe, and Proliferative DR~\cite{wilkinson2003proposed} and early detection and appropriate management are important for preventing vision-threatening progression~\cite{rohan1989prevention}. Retinal fundus photography offers a non-invasive and scalable screening modality, and deep-learning models trained on large fundus-image datasets have demonstrated diagnostic precision similar to that of specialist ophthalmologists~\cite{gulshan2016development,lam2018automated}.

Despite this promise, deploying high-capacity architectures such as ResNet-50~\cite{he2016deep} and Swin Transformer~\cite{liu2021swin} on resource-constrained devices remains challenging due to their relatively high computational and memory requirements. This limitation is particularly relevant for community and remote screening, where AI systems may need to operate on mobile or edge devices with limited processing power and connectivity~\cite{joshi2025mobile}. Lightweight convolutional networks, including the MobileNetV3 family, offer a practical alternative for resource-constrained deployment~\cite{howard2019searching}. However, their reduced model capacity can introduce an accuracy--efficiency trade-off; in a recent diabetic retinopathy study, MobileNetV3 variants achieved lower classification accuracy than ResNet-50 while using substantially fewer parameters and providing considerably higher inference speed~\cite{zhang2025dual}.

Knowledge distillation (KD) provides a principled approach for transferring knowledge from high-capacity models to more compact networks. In conventional offline KD, a high-capacity \emph{teacher} network is first trained, after which a compact \emph{student} is optimised using both the teacher's temperature-scaled soft predictions and the ground-truth labels. The resulting soft targets provide information about inter-class similarities that is absent from one-hot supervision, offering additional supervisory signals that can regularise student training~\cite{hinton2015distilling}. While the effectiveness of KD for general image classification has been extensively demonstrated~\cite{hinton2015distilling,beyer2022goodteacher}, its effectiveness can vary with the target task and teacher--student configuration. In particular, the application of KD to ordinal diabetic retinopathy grading, where neighbouring severity levels have meaningful clinical relationships, remains comparatively less explored.

In this paper, we address three concrete questions. First, how does the choice of teacher architecture, spanning a classical convolutional neural network (CNN) (ResNet-50), a lightweight CNN (MobileNetV3-Large), and a hierarchical vision transformer (Swin Transformer-Tiny, Swin-T), affect the quality of knowledge transferred to a fixed MobileNetV3-Small student? Second, how sensitive is distillation performance to the softening temperature $T$ and the hard-label weighting $\alpha$ for each teacher--student pair, and can a single optimal temperature--weighting combination be recommended across all pairings? Third, does distillation yield clinically meaningful per-class improvements, particularly for the minority and ambiguous severity grades?

To answer these questions, we train and evaluate all teacher--student combinations on the APTOS 2019 Blindness Detection dataset under a unified experimental protocol. We conduct a temperature ablation over $T \in \{2, 3, 4, 5\}$ for each pairing and additionally evaluate the effect of the hard-label weighting $\alpha$ across selected configurations. Our main findings are as follows:

\begin{itemize}
\item \textbf{Overall KD performance.}
Across the three teacher architectures and four distillation temperatures, eight of the twelve single-seed KD configurations outperform the MobileNetV3-Small baseline. ResNet-50 and MobileNetV3-Large provide the most consistent improvements, while Swin-T achieves the highest single-run quadratic weighted kappa (QWK) of $0.9093$ at $T{=}4$, $\alpha{=}0.7$, with a macro-F1 of $0.6851$.

\item \textbf{Sensitivity to distillation hyperparameters.}
KD performance varies substantially with both temperature $T$ and hard-label weighting $\alpha$, with the optimal temperature differing across teacher architectures: $T{=}2$ for MobileNetV3-Large, $T{=}4$ for Swin-T, and $T{=}5$ for ResNet-50. The $\alpha$ ablation similarly shows configuration-dependent effects, with $\alpha{=}0.3$ favoured at the tested temperature extremes and $\alpha{=}0.7$ generally favoured at $T{=}4$.

\item \textbf{Deployment efficiency.}
The distilled MobileNetV3-Small retains a compact 5.94~MB model footprint. This combination of competitive predictive performance and low computational overhead supports the feasibility of deploying the model for resource-constrained or mobile DR screening.
\end{itemize}

\section{Related work}

\subsection{Deep learning for diabetic retinopathy grading}

The use of deep learning for DR detection has evolved over the past decade. Earlier approaches, such as the DREAM system, used handcrafted retinal features with traditional machine-learning classifiers~\cite{roychowdhury2014dream}. While effective, these methods relied on task-specific feature engineering, motivating the shift toward deep learning for automated feature extraction. Around 2015, CNNs entered the field, initially targeting narrow tasks like microaneurysm detection for early diabetic retinopathy screening~\cite{haloi2015improved}. A major turning point came with the release of a large, publicly labelled fundus dataset through a Kaggle competition~\cite{kaggle2015}, enabling proper training of deep networks. Pratt et al. then developed an early CNN-based approach for five-stage DR classification, incorporating data augmentation and training on the publicly available Kaggle fundus-image dataset~\cite{pratt2016convolutional}. Soon after, Gulshan et al. demonstrated that a deep learning algorithm could achieve high sensitivity and specificity for referable DR, with performance evaluated against expert ophthalmologist grading~\cite{gulshan2016development}. Parallel work integrated deep learning into an existing automated DR detection pipeline, demonstrating improved performance on a publicly available dataset~\cite{abramoff2016improved}. By 2017, research expanded toward broader validation, including detailed disease staging~\cite{takahashi2017applying} and multiethnic, multi-disease screening systems~\cite{ting2017development}. Attention then shifted to interpretability, with methods for precise lesion localisation and visual explanation of model decisions~\cite{orlando2018ensemble,wang2018diabetic}. Recent work addresses data scarcity and grader inconsistency through transfer learning and ordinal severity modelling, achieving strong results on benchmarks like APTOS 2019.

\subsection{Knowledge distillation}

KD was first introduced as a general model-compression technique and has since been widely adopted across many areas of deep learning, including natural language processing, speech, and computer vision, wherever transferring the behaviour of a large, accurate model into a smaller, faster one is valuable. In vision specifically, KD was originally proposed as a way to train a compact ``student'' network to mimic the softened output probabilities of a larger, pretrained ``teacher'' network, capturing richer inter-class similarity information than hard labels alone~\cite{hinton2015distilling}. Soon after, researchers extended this idea beyond output matching to intermediate representations: FitNets introduced the idea of using a teacher's mid-network feature activations as ``hints'' to guide the student's learning, allowing thinner networks to be compressed while retaining much of the teacher's accuracy~\cite{romero2014fitnets}. This was followed by attention transfer methods, which aligned the spatial attention maps of teacher and student networks at multiple points rather than relying on a single hint location, further improving feature-level knowledge transfer in CNNs~\cite{zagoruyko2016paying}. Later work broadened KD beyond simple classification to more complex vision tasks such as object detection and segmentation, and introduced relational and contrastive variants that preserve similarity structures between samples rather than matching individual outputs or features~\cite{park2019relational,tian2019contrastive}. With the rise of vision transformers, KD was adapted to help data-hungry transformer architectures train efficiently with substantially less pretraining data, most notably through a dedicated distillation token that let a transformer learn directly from a CNN teacher's inductive biases~\cite{touvron2021training}. These developments have established KD as an important model-compression strategy for transferring the capabilities of computationally expensive models to more efficient architectures, particularly when deploying deep learning models under computational and memory constraints.

\subsection{Knowledge distillation for DR classification}

Several studies have applied KD specifically to DR classification. Abbasi et al. propose a semi-supervised KD approach that exploits unlabelled fundus images to augment a teacher--student training pipeline on the Messidor and EyePACS datasets, demonstrating that incorporating unlabelled data improves distillation quality when annotations are scarce~\cite{abbasi2021classification}. Moya-Albor et al. present a computer-aided diagnosis system using an Inception-v3 teacher and an Inception-v3 student with a combined Kullback--Leibler (KL) divergence and categorical cross-entropy loss to address class imbalance in DR lesion datasets~\cite{math12162543}. Islam et al. go further by fusing a ResNet-152 and a Swin Transformer into a composite teacher that distils into a compact Xception-CBAM student, training on both APTOS 2019 and IDRiD~\cite{app132212397}. Shafi et al. use a transformer-based teacher (MobileViTv2) to distil into GoogLeNet with Grad-CAM explanations on the APTOS dataset~\cite{shafi2024ijcnn}. Dejene et al. specifically target a lightweight MobileNet student on APTOS 2019, reporting 98.38\% binary-classification accuracy and demonstrating the feasibility of KD for low-resource DR screening~\cite{dejene2025knowledge}. Wang et al. extend the KD paradigm to DR \emph{segmentation}, proposing a lesion-aware distillation framework with a lesion embedding queue on IDRiD and DDR~\cite{wang2024lesion}.

While these works collectively establish KD as a viable strategy for DR modelling, they vary widely in architecture choices, datasets, and evaluation protocols, and none systematically compares CNN-based and transformer-based teachers on the same student across a range of temperatures. The present work addresses this gap directly.

\section{Materials and methods}

The workflow proceeds in four stages: preprocessing, teacher training, knowledge distillation, and evaluation.

\subsection{Dataset}

All experiments use the Asia Pacific Tele-Ophthalmology Society (APTOS) 2019 Blindness Detection dataset~\cite{aptos2019}, which contains 3,662 retinal fundus photographs collected at Aravind Eye Hospital, India, and graded by trained clinicians into five DR severity classes: 0~=~No DR, 1~=~Mild, 2~=~Moderate, 3~=~Severe, and 4~=~Proliferative DR.

We did not create our own data partition. We used a publicly available pre-split version of the dataset hosted on Kaggle~\cite{aptos_split}, which divides the 3,662 labelled images into training (2,930), validation (366) and test (366) subsets. The partitions were used exactly as provided, without re-splitting or modification. Table~\ref{tab:dataset} reports the per-class counts computed from the provided label files. To our knowledge, the splitting procedure, including whether it was stratified and the random seed used, is not documented by the dataset's author, and the class proportions in the test subset differ from those of the full dataset (for example, No DR accounts for 54\% of the test subset versus 49\% overall). The random seed (42) reported elsewhere in this paper controls model initialisation and training only, not the data partition. APTOS 2019 does not distribute patient identifiers, so we could not verify that images from the same patient are confined to a single subset; this is noted as a limitation. This study used only publicly available, de-identified retinal images; no new data were collected from human participants, and no additional ethical approval or consent was required.

\begin{table}[!htbp]
\centering
\caption{\textbf{APTOS 2019 dataset statistics in the provided training, validation and test subsets.} Counts reflect the number of images per class and split used in all experiments.}
\begin{tabular}{|l|r|r|r|r|r|}
\hline
\textbf{Split} & \textbf{No DR} & \textbf{Mild} & \textbf{Moderate} & \textbf{Severe} & \textbf{Prolif.} \\ \hline
Train (80\%) & 1,434 & 300 & 808 & 154 & 234 \\ \hline
Val (10\%)   &   172 &  40 & 104 &  22 &  28 \\ \hline
Test (10\%)  &   199 &  30 &  87 &  17 &  33 \\ \hline
Total        & 1,805 & 370 & 999 & 193 & 295 \\ \hline
\end{tabular}
\label{tab:dataset}
\end{table}

The dataset exhibits substantial class imbalance: No-DR images account for roughly 50\% of the total, while Severe DR constitutes only about 5\%. To mitigate this, we apply inverse-frequency class weighting to the teacher's cross-entropy loss and use a \texttt{WeightedRandomSampler} to oversample minority classes.

\subsection{Preprocessing}

All images are resized to $224 \times 224$ pixels to satisfy the input requirements of all three teacher backbones and normalised using the ImageNet channel statistics (mean $= [0.485, 0.456, 0.406]$, std $= [0.229, 0.224, 0.225]$), consistent with the pretrained weight initialisation.

Training augmentation differs between the teacher and student pipelines. The teacher is trained with a heavier augmentation regime that includes random horizontal and vertical flips (each $p = 0.5$), rotation up to $20^{\circ}$, colour jitter (brightness and contrast $\pm 0.3$, saturation $\pm 0.2$, hue $\pm 0.05$), and random affine perturbation ($\pm 5\%$ translation and scaling). These augmentations encourage the teacher to learn representations that are robust to the photometric variability commonly encountered in real-world fundus images. The student is trained with a lighter regime consisting of horizontal and vertical flips, rotation up to $15^{\circ}$, and modest colour jitter (brightness and contrast $\pm 0.2$ only), without affine transformation, saturation, or hue perturbation.

Saturation and hue augmentation are deliberately excluded from the student pipeline because colour is diagnostically meaningful in retinal fundus images. In particular, the visibility of microaneurysms and the appearance of retinal vessels can depend on colour information. Excessive colour perturbation may therefore encourage the student to learn unnecessary colour invariance. In addition, the student gains some robustness to photometric variation through the teacher's soft-label distributions, which capture the teacher's responses to diverse augmented inputs without requiring the student to experience the full range of augmentation directly~\cite{hinton2015distilling}.

During distillation, the frozen teacher and the student process identical image tensors drawn from the student training loader. Thus, the soft targets represent the teacher's response to the same lightly augmented inputs used by the student.

\subsection{Teacher training}

Three teacher architectures are evaluated in this study: ResNet-50, MobileNetV3-Large, and Swin-Transformer-Tiny (hereafter Swin-T). Each is initialised from ImageNet pretrained weights and trained end-to-end on the APTOS training set, with only the classification head modified to produce five output logits corresponding to the DR severity grades.

\subsubsection{ResNet-50}
The default fully-connected head is replaced with a single linear layer mapping the 2048-dimensional penultimate feature vector directly to five class logits.

\subsubsection{MobileNetV3-Large}
The final linear layer of the four-layer classifier block is replaced with a linear layer producing five logits. All preceding layers of the classifier are retained and trained end-to-end alongside the backbone.

\subsubsection{Swin-T}
The default classification head is replaced with a single linear layer projecting the 768-dimensional output representation to five class logits.

\subsubsection{Training configuration}
ResNet-50 and MobileNetV3-Large share an identical training configuration: the Adam optimiser at a learning rate of $10^{-3}$ and weight decay $10^{-4}$. Swin-T is trained with AdamW~\cite{loshchilov2017decoupled} at a lower learning rate of $10^{-4}$, which is standard practice for vision transformer fine-tuning, where larger gradient steps risk destabilising the self-attention layers during adaptation to a new domain~\cite{beyer2022goodteacher}. All three teachers use a cosine annealing learning-rate schedule (minimum $10^{-6}$), class-weighted cross-entropy loss to address the pronounced class imbalance in APTOS~2019, and a \texttt{WeightedRandomSampler} to ensure balanced class representation within each mini-batch. Training runs for 30 epochs with a batch size of 32, and the checkpoint that achieves the highest validation QWK is selected. Following training, the teacher is set to evaluation mode and all its parameters are frozen permanently; no teacher gradient is computed at any point during student training.

\subsection{Knowledge distillation}

We follow the response-based offline KD formulation of Hinton et al.~\cite{hinton2015distilling}. Given a batch of images $\mathbf{x}$, let $\mathbf{z}^T$ and $\mathbf{z}^S$ denote the logits produced by the frozen teacher and the trainable student, respectively, and let $\mathbf{y}$ denote the hard ground-truth labels. The total training loss is:

\begin{equation}
  \mathcal{L} \;=\;
    \alpha \cdot \mathcal{L}_{\text{CE}}
      (\mathbf{z}^S,\,\mathbf{y})
    \;+\;
    (1 - \alpha) \cdot T^{2} \cdot
    \mathcal{L}_{\text{KL}}\!\left(
      \sigma\!\left(\tfrac{\mathbf{z}^T}{T}\right)
      \,\Big\|\,
      \sigma\!\left(\tfrac{\mathbf{z}^S}{T}\right)
    \right),
  \label{eq:kd_loss}
\end{equation}

where $\sigma(\cdot)$ denotes the softmax function, $\mathcal{L}_{\text{CE}}$ is unweighted cross-entropy, $\mathcal{L}_{\text{KL}}$ is the KL divergence with \texttt{batchmean} reduction, $\alpha$ is the hard-label weight, and $T$ is the distillation temperature. The factor $T^{2}$ compensates for the reduction in gradient magnitude that occurs as $T$ increases, keeping the contribution of the soft-target term commensurate with the cross-entropy term across the temperature range tested~\cite{hinton2015distilling}. In the primary experiments $\alpha$ is set to 0.3, placing 70\% of the training signal on the teacher's soft distributions; a secondary ablation over $\alpha \in \{0.3, 0.7\}$ is reported in the Results (Alpha sensitivity).

\subsubsection{Student architecture}
The student is MobileNetV3-Small, initialised from ImageNet pretrained weights. Only the final linear layer of the four-layer classifier head is replaced with a linear mapping to five output logits; the preceding layers (Linear(576, 1024), Hardswish, Dropout(0.2)) are retained from the pretrained checkpoint.

\subsubsection{Student training protocol}
The student is trained with the Adam optimiser at learning rate $10^{-3}$ and weight decay $10^{-4}$, with a cosine annealing schedule (minimum $10^{-6}$) and a \texttt{WeightedRandomSampler} to maintain balanced class representation within each mini-batch. Training runs for 30 epochs with batch size 32. Validation performance is monitored using unweighted cross-entropy loss, and the checkpoint achieving the highest validation QWK is selected for final evaluation.

\subsection{Temperature ablation}

To identify the optimal distillation temperature for each teacher, we sweep $T \in \{2, 3, 4, 5\}$ with $\alpha$ fixed at 0.3. For each value of $T$, a fresh student is initialised from the same ImageNet-pretrained checkpoint to eliminate order-dependent variance between runs. Each teacher is trained once and frozen before the sweep begins, so that any difference in student performance across temperatures is attributable solely to $T$.

\subsection{Alpha sensitivity ablation}

To determine whether the relative weighting of soft-target and hard-label supervision depends on teacher architecture, temperature, or their interaction, the temperature ablation described above is extended with a second hard-label weight, $\alpha{=}0.7$, applied across the same temperature range $T \in \{2,3,4,5\}$ and the same three teacher--student pairings. Combined with the temperature ablation, this yields a $3 \times 4 \times 2$ factorial design (teacher $\times$ temperature $\times$ $\alpha$), for 24 configurations in total. As in the temperature ablation, a fresh student is initialised from the same ImageNet-pretrained checkpoint for every configuration, and each teacher checkpoint from the Teacher training subsection is reused unchanged across all $(T, \alpha)$ combinations for that teacher.

\subsection{Multi-seed robustness verification}

The primary ablations were conducted with a single random seed (seed~42). Two distinct patterns in the results motivated a targeted multi-seed follow-up. First, the ResNet-50 pipeline at $T{=}4$ produced seed-42 test QWK values (0.8666 and 0.8698) below both the no-KD baseline and the same pipeline at adjacent temperatures, with validation--test QWK gaps of 0.0321 and 0.0314. This pattern is inconsistent with a genuine temperature effect and suggests a seed-specific artefact. Second, at ResNet-50 $T{=}3$ and Swin-T $T{=}4$, the single-seed margin between $\alpha{=}0.3$ and $\alpha{=}0.7$ was narrow enough (below 0.003 QWK) that the direction of the difference could plausibly reverse under a different initialisation.

To address both concerns, two additional random seeds (123, 999) were used to retrain the student at ResNet-50 $T{=}3$, ResNet-50 $T{=}4$, and Swin-T $T{=}4$, for both $\alpha$ values at each configuration. This yields three seeds in total per (teacher, $T$, $\alpha$) triplet, covering six repeated configurations overall. All hyperparameters other than the random seed follow the Knowledge distillation subsection.

For transparency, a run was flagged when its validation--test QWK gap was approximately seven times the mean gap of the other two seeds at the same configuration. Flagged runs are reported individually and retained in all primary means; a sensitivity analysis excluding them is also reported. The flagging rule was defined after inspecting the seed-42 results and is descriptive, not a pre-specified exclusion criterion.

\subsection{Baseline}

To quantify the benefit of distillation, a MobileNetV3-Small baseline is trained under the same student protocol described in the Knowledge distillation subsection but with no teacher. The loss reduces to standard unweighted cross-entropy on hard labels, with all other hyperparameters held constant.

\subsection{Evaluation protocol}

All models are evaluated on the same held-out test set. The primary metric is \textbf{quadratic weighted kappa (QWK)}, which penalises misclassifications proportionally to the squared distance between the predicted and true ordinal grade and is the standard metric for DR severity assessment. Secondary metrics include top-1 accuracy, macro-averaged F1 score (F1$_\text{m}$), and per-class F1. Inference efficiency is characterised by parameter count, model size (MB), and single-image latency (mean of 100 GPU forward passes). The performance retention metric:

\begin{equation}
  \rho = \frac{\text{QWK}_{\text{student}}}
              {\text{QWK}_{\text{teacher}}} \times 100\%
  \label{eq:retention}
\end{equation}

quantifies what fraction of the teacher's discriminative ability is recovered in the compressed student.

Table~\ref{tab:hyperparams} summarises all fixed hyperparameters used throughout the study.

\begin{table}[!htbp]
\centering
\caption{\textbf{Fixed hyperparameters shared across all experiments.}}
\begin{tabular}{|l|l|}
\hline
\textbf{Hyperparameter} & \textbf{Value} \\ \hline
Image size              & $224 \times 224$ \\ \hline
Batch size              & 32 \\ \hline
Epochs                  & 30 \\ \hline
Optimiser (ResNet-50, MobileNetV3-Large, student) & Adam \\ \hline
Optimiser (Swin-T teacher)                        & AdamW \\ \hline
Learning rate (ResNet-50, MobileNetV3-Large, student) & $1 \times 10^{-3}$ \\ \hline
Learning rate (Swin-T teacher)             & $1 \times 10^{-4}$ \\ \hline
LR schedule             & Cosine annealing (min $10^{-6}$) \\ \hline
Weight decay            & $1 \times 10^{-4}$ \\ \hline
KD weight $\alpha$ (temperature ablation) & $0.3$ \\ \hline
KD weight $\alpha$ (alpha ablation) & $\{0.3,\;0.7\}$, all 3 teachers \\ \hline
Temperature $T$         & $\{2, 3, 4, 5\}$, all 3 teachers \\ \hline
Random seed (primary)   & 42 \\ \hline
Random seed (repeated-seed verification) & 123, 999 (6 configurations) \\ \hline
\end{tabular}
\label{tab:hyperparams}
\end{table}

\subsection{Implementation and code availability}

All models were implemented in Python (version 3.12.12) using PyTorch (version 2.10.0) and torchvision (version 0.25.0). Quadratic weighted kappa (QWK) and macro-averaged F1-score were computed using scikit-learn (version 1.6.1). Experiments, including latency measurements, were conducted in the Kaggle Notebooks environment. The code, experimental notebooks, and associated research materials are available from Zenodo at \url{https://doi.org/10.5281/zenodo.23533877}.

\section{Results}

This section reports, in order: test-set performance of the three teacher checkpoints; a three-seed no-KD MobileNetV3-Small baseline; a single-seed temperature ablation at $\alpha=0.3$; a single-seed alpha-sensitivity analysis at $\alpha \in \{0.3,0.7\}$; repeated-seed verification of six selected configurations; and per-class and efficiency analyses of selected student configurations. All models were evaluated on the same held-out APTOS 2019 test set. QWK served as the primary metric, while accuracy and macro-F1 were reported as secondary metrics. Within each run, the checkpoint with the highest validation QWK was selected, and the selected checkpoint was then evaluated once on the test set.

Two reporting conventions follow from the study design. First, the temperature--alpha configurations were compared using their test-set results; therefore, the highest observed value across the grid is reported as a screening outcome rather than as an unbiased estimate of the selected configuration's generalisation performance. Second, all 24 grid configurations were initially evaluated using a single seed, while six selected configurations were subsequently repeated using three seeds. Single-seed and multi-seed results are reported separately and are not pooled.

\subsection{Teacher model performance}

Table~\ref{tab:teacher} reports test-set performance and computational characteristics for the three teacher architectures, each trained under the protocol described in the Teacher training subsection of the Materials and methods.

Table~\ref{tab:teacher_perclass} reports per-class F1 for each teacher.

\begin{table}[!htbp]
\centering
\caption{\textbf{Test-set performance and efficiency of the three teacher architectures.}}
\begin{tabular}{|l|r|r|r|r|r|r|}
\hline
\textbf{Teacher} & \textbf{Params (M)} & \textbf{Size (MB)} & \textbf{Lat. (ms)} & \textbf{QWK} & \textbf{Acc.} & \textbf{macro-F1} \\ \hline
ResNet-50   & 23.52 &  90.02 &  5.61 & 0.8793 & 0.8169 & 0.6589 \\ \hline
MobileNetV3-Large &  4.21 &  16.25 &  6.33 & 0.9084 & 0.8333 & 0.6449 \\ \hline
Swin-T     & 27.52 & 105.28 & 12.22 & 0.9024 & 0.8197 & 0.6540 \\ \hline
\end{tabular}
\label{tab:teacher}
\end{table}

\begin{table}[!htbp]
\centering
\caption{\textbf{Teacher per-class F1 on the test set.}}
\begin{tabular}{|l|r|r|r|r|r|}
\hline
\textbf{Teacher} & \textbf{No DR} & \textbf{Mild} & \textbf{Moderate} & \textbf{Severe} & \textbf{Prolif.} \\ \hline
ResNet-50   & 0.9823 & 0.6176 & 0.7000 & 0.4186 & 0.5758 \\ \hline
MobileNetV3-Large & 0.9800 & 0.5538 & 0.7709 & 0.3125 & 0.6071 \\ \hline
Swin-T     & 0.9847 & 0.6111 & 0.7284 & 0.3556 & 0.5902 \\ \hline
\end{tabular}
\label{tab:teacher_perclass}
\end{table}

\subsection{Student baseline (no KD)}

The MobileNetV3-Small student was trained without a teacher under the baseline protocol described in the Materials and methods at three random seeds (42, 123, 999). Table~\ref{tab:baseline_seeds} reports test-set metrics for each seed, and Table~\ref{tab:baseline_perclass} the corresponding per-class F1 scores.

\begin{table}[!htbp]
\centering
\caption{\textbf{No-KD MobileNetV3-Small baseline across three random seeds, held-out test set.}}
\begin{tabular}{|c|r|r|r|}
\hline
\textbf{Seed} & \textbf{Test QWK} & \textbf{Test Acc.} & \textbf{macro-F1} \\ \hline
42  & 0.8913 & 0.8169 & 0.6190 \\ \hline
123 & 0.8948 & 0.8361 & 0.6522 \\ \hline
999 & 0.8551 & 0.8033 & 0.5871 \\ \hline
\textbf{Mean}  & \textbf{0.8804} & 0.8188 & 0.6194 \\ \hline
\textbf{Range} & 0.0397 & 0.0328 & 0.0651 \\ \hline
\end{tabular}
\label{tab:baseline_seeds}
\end{table}

\begin{table}[!htbp]
\centering
\caption{\textbf{No-KD baseline per-class F1 across three random seeds.}}
\begin{tabular}{|c|r|r|r|r|r|}
\hline
\textbf{Seed} & \textbf{No DR} & \textbf{Mild} & \textbf{Moderate} & \textbf{Severe} & \textbf{Prolif.} \\ \hline
42  & 0.9823 & 0.5000 & 0.7407 & 0.3333 & 0.5385 \\ \hline
123 & 0.9799 & 0.5357 & 0.7849 & 0.3784 & 0.5818 \\ \hline
999 & 0.9626 & 0.4400 & 0.7411 & 0.3429 & 0.4490 \\ \hline
\textbf{Mean} & 0.9749 & 0.4919 & 0.7556 & 0.3515 & 0.5231 \\ \hline
\end{tabular}
\label{tab:baseline_perclass}
\end{table}

Two reference values follow from Table~\ref{tab:baseline_seeds} and are used consistently below. Seed 42 (QWK 0.8913) is the \emph{single-seed reference baseline} for the temperature and alpha-screening tables because those configurations were trained at seed 42. The three-seed mean (QWK 0.8804) is reported as the \emph{multi-seed reference baseline} when interpreting repeated-seed distillation results.

Baseline QWK ranged from 0.8551 to 0.8948 across the three seeds. This seed-dependent variation is considered when interpreting the single-seed screening results; repeated-seed comparisons are reported in the Multi-seed robustness verification subsection below.

\subsection{Ablation with \texorpdfstring{$\alpha = 0.3$}{alpha = 0.3}}

Tables~\ref{tab:ablation_resnet}, \ref{tab:ablation_mnv3}, and~\ref{tab:ablation_swint} report validation and test performance across $T \in \{2,3,4,5\}$ for each teacher at the primary hard-label weight, $\alpha{=}0.3$. For every configuration, the checkpoint with the highest validation QWK was selected, and the selected checkpoint was then evaluated once on the held-out test set.

\begin{table}[!htbp]
\centering
\caption{\textbf{Temperature ablation, ResNet-50$\to$MobileNetV3-Small ($\alpha{=}0.3$).}}
\begin{tabular}{|c|r|r|r|r|r|r|}
\hline
\textbf{$T$} & \textbf{Val QWK} & \textbf{Test QWK} & \textbf{Test Acc.} & \textbf{macro-F1} & \textbf{Ret.\ (\%)} & \textbf{Best Ep.} \\ \hline
2 & 0.8977 & 0.9019 & 0.8224 & 0.6539 & 102.6 & 22 \\ \hline
3 & 0.9048 & 0.8939 & 0.8142 & 0.6425 & 101.7 & 22 \\ \hline
4 & 0.8987 & 0.8666 & 0.7814 & 0.5965 &  98.6 & 12 \\ \hline
5 & 0.9050 & 0.9090 & 0.8142 & 0.6537 & 103.4 & 24 \\ \hline
\end{tabular}
\begin{flushleft} Teacher test QWK: 0.8793. Reference baseline (seed 42) test QWK: 0.8913.
\end{flushleft}
\label{tab:ablation_resnet}
\end{table}

\begin{table}[!htbp]
\centering
\caption{\textbf{Temperature ablation, MobileNetV3-Large$\to$MobileNetV3-Small ($\alpha{=}0.3$).}}
\begin{tabular}{|c|r|r|r|r|r|r|}
\hline
\textbf{$T$} & \textbf{Val QWK} & \textbf{Test QWK} & \textbf{Test Acc.} & \textbf{macro-F1} & \textbf{Ret.\ (\%)} & \textbf{Best Ep.} \\ \hline
2 & 0.9190 & 0.9066 & 0.8443 & 0.6695 & 99.8 & 27 \\ \hline
3 & 0.8988 & 0.8932 & 0.8333 & 0.6418 & 98.3 & 20 \\ \hline
4 & 0.8986 & 0.8822 & 0.8005 & 0.5993 & 97.1 & 22 \\ \hline
5 & 0.9041 & 0.8919 & 0.8251 & 0.6299 & 98.2 & 20 \\ \hline
\end{tabular}
\begin{flushleft} Teacher test QWK: 0.9084. Reference baseline (seed 42) test QWK: 0.8913.
\end{flushleft}
\label{tab:ablation_mnv3}
\end{table}

\begin{table}[!htbp]
\centering
\caption{\textbf{Temperature ablation, Swin-T$\to$MobileNetV3-Small ($\alpha{=}0.3$).}}
\begin{tabular}{|c|r|r|r|r|r|r|}
\hline
\textbf{$T$} & \textbf{Val QWK} & \textbf{Test QWK} & \textbf{Test Acc.} & \textbf{macro-F1} & \textbf{Ret.\ (\%)} & \textbf{Best Ep.} \\ \hline
2 & 0.9064 & 0.8909 & 0.8197 & 0.6560 & 98.7 & 22 \\ \hline
3 & 0.9010 & 0.8762 & 0.7978 & 0.6081 & 97.1 & 13 \\ \hline
4 & 0.9050 & 0.9065 & 0.8251 & 0.6620 & 100.5 & 28 \\ \hline
5 & 0.9115 & 0.9008 & 0.8169 & 0.6407 & 99.8 & 22 \\ \hline
\end{tabular}
\begin{flushleft} Teacher test QWK: 0.9024. Reference baseline (seed 42) test QWK: 0.8913.
\end{flushleft}
\label{tab:ablation_swint}
\end{table}

Across the twelve evaluated configurations, eight produced a higher observed test QWK than the no-KD baseline (0.8913), while four produced lower values: ResNet-50 at $T=4$ (0.8666), MobileNetV3-Large at $T=4$ (0.8822), and Swin-T at $T=2$ (0.8909) and $T=3$ (0.8762). The highest observed test QWK for each teacher was 0.9090 for ResNet-50 at $T=5$, 0.9066 for MobileNetV3-Large at $T=2$, and 0.9065 for Swin-T at $T=4$.

These are single-seed screening results. The held-out test values are reported for comparative analysis across the planned temperature sweep and should not be interpreted as independent estimates after selecting the highest observed configuration.

\subsection{Alpha sensitivity \texorpdfstring{$\alpha \in \{0.3, 0.7\}$}{alpha in {0.3, 0.7}}}

Table~\ref{tab:alpha_grid} shows a clear temperature-linked pattern in the sign of the $\alpha$ preference. At the temperature extremes tested, $\alpha{=}0.3$ yielded the higher observed test QWK for all three teachers ($T{=}2$: $-$0.0144, $-$0.0197, $-$0.0127; $T{=}5$: $-$0.0148, $-$0.0082, $-$0.0082), whereas at $T{=}4$ all three teachers favoured $\alpha{=}0.7$ ($+$0.0032, $+$0.0195, $+$0.0028). $T{=}3$ was the only temperature at which the teachers disagreed in direction. Under the $\alpha{=}0.7$ arm the temperature associated with each teacher's highest observed test QWK was $T{=}5$ for ResNet-50 (0.8942), $T{=}4$ for MobileNetV3-Large (0.9017) and $T{=}4$ for Swin-T (0.9093); MobileNetV3-Large's optimum thus shifted relative to the $\alpha{=}0.3$ arm, suggesting an interaction between the two hyperparameters in the reported screening results.

\begin{table}[!htbp]
\centering
\caption{\textbf{Test QWK by temperature and alpha, all three teachers.}}
\begin{tabular}{|l|c|r|r|r|}
\hline
\textbf{Teacher} & \textbf{$T$} & \textbf{$\alpha{=}0.3$} & \textbf{$\alpha{=}0.7$} & \textbf{$\Delta$} \\ \hline
ResNet-50   & 2 & 0.9019 & 0.8875 & $-$0.0144 \\ \hline
ResNet-50   & 3 & 0.8939 & 0.8922 & $-$0.0017 \\ \hline
ResNet-50   & 4 & 0.8666 & 0.8698 & $+$0.0032 \\ \hline
ResNet-50   & 5 & 0.9090 & 0.8942 & $-$0.0148 \\ \hline
MobileNetV3-Large & 2 & 0.9066 & 0.8869 & $-$0.0197 \\ \hline
MobileNetV3-Large & 3 & 0.8932 & 0.8672 & $-$0.0260 \\ \hline
MobileNetV3-Large & 4 & 0.8822 & 0.9017 & $+$0.0195 \\ \hline
MobileNetV3-Large & 5 & 0.8919 & 0.8837 & $-$0.0082 \\ \hline
Swin-T     & 2 & 0.8909 & 0.8782 & $-$0.0127 \\ \hline
Swin-T     & 3 & 0.8762 & 0.8924 & $+$0.0162 \\ \hline
Swin-T     & 4 & 0.9065 & 0.9093 & $+$0.0028 \\ \hline
Swin-T     & 5 & 0.9008 & 0.8926 & $-$0.0082 \\ \hline
\end{tabular}
\begin{flushleft} $\Delta$ is test QWK at $\alpha{=}0.7$ minus test QWK at $\alpha{=}0.3$.
\end{flushleft}
\label{tab:alpha_grid}
\end{table}

\begin{figure}[!htbp]
\centering
\includegraphics[width=0.85\linewidth]{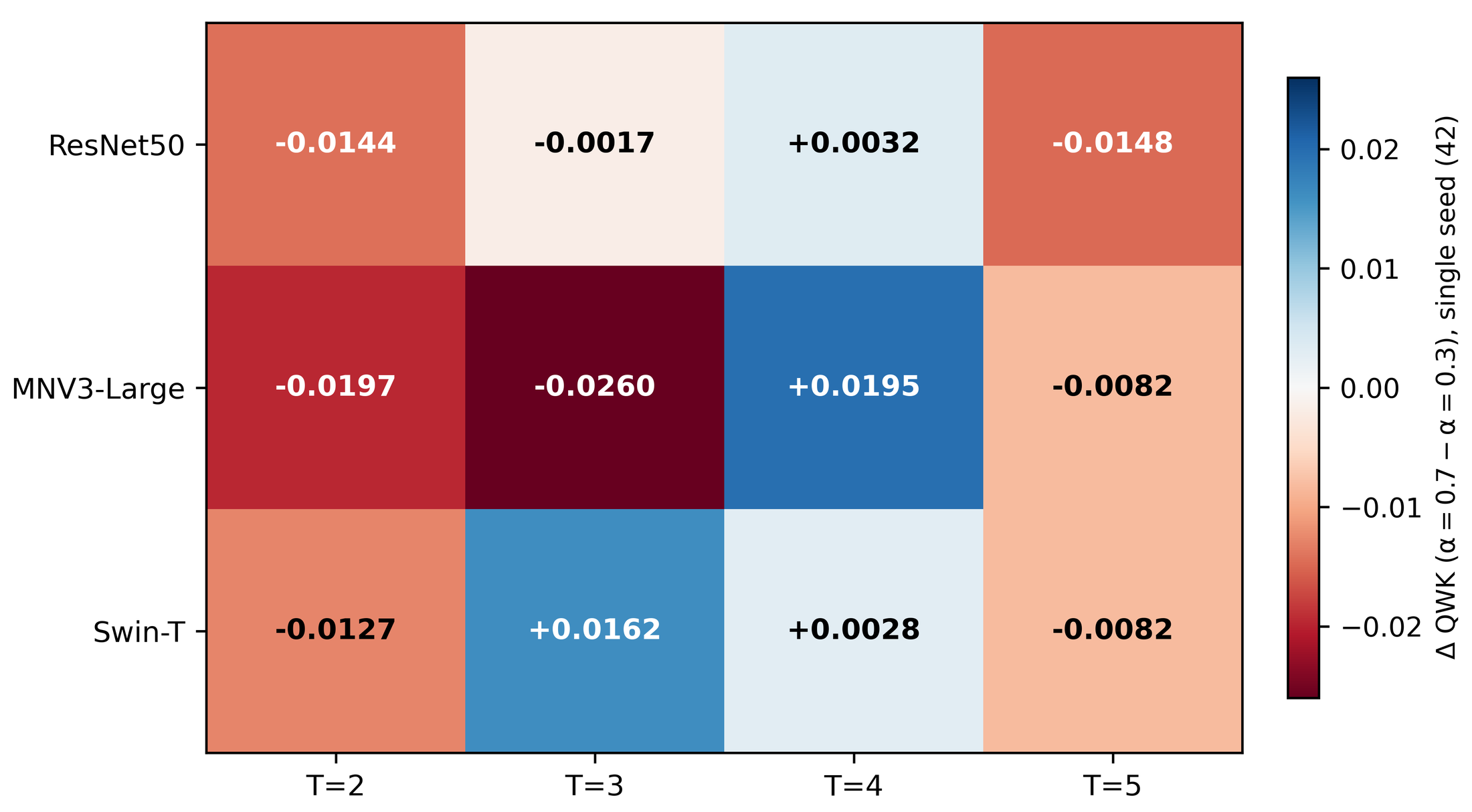}
\caption{\textbf{Effect of the hard-label weight $\alpha$ on test QWK across the single-seed screening grid.}
Heat map of $\Delta$ (test QWK at $\alpha{=}0.7$ minus test QWK at $\alpha{=}0.3$) for each teacher and temperature $T$, seed 42 only (values as in Table~\ref{tab:alpha_grid}). Blue cells indicate a higher QWK at $\alpha{=}0.7$ and red cells a higher QWK at $\alpha{=}0.3$. Repeated-seed results for selected configurations are shown in Fig~\ref{fig:seed_verification}.}
\label{fig:alpha_heatmap}
\end{figure}

Fig~\ref{fig:alpha_heatmap} visualises the $\Delta$ column of Table~\ref{tab:alpha_grid} as a heat map, showing the sign of the $\alpha$ preference by teacher and temperature.

All twelve pairs were initially screened at seed 42. The three configurations with the smallest observed $|\Delta|$ values were selected for repeated-seed verification: ResNet-50 at $T{=}3$ (0.0017), Swin-T at $T{=}4$ (0.0028), and ResNet-50 at $T{=}4$ (0.0032). These values were all below 0.004 QWK. The selection was motivated by the possibility that a small single-seed margin could change under a different initialisation.

Validation and test QWK and additional metrics (Acc., F1$_\text{m}$, retention) for each of the twelve $\alpha{=}0.7$ configurations above are reported in Table~\ref{tab:alpha_full}.

\begin{table}[!htbp]
\centering
\caption{\textbf{Alpha $=$ 0.7 sweep, full metrics, all three teachers.}}
\begin{tabular}{|l|c|r|r|r|r|r|}
\hline
\textbf{Teacher} & \textbf{$T$} & \textbf{Val QWK} & \textbf{Test QWK} & \textbf{Acc.} & \textbf{macro-F1} & \textbf{Ret.\ (\%)} \\ \hline
ResNet-50   & 2 & 0.9172 & 0.8875 & 0.8388 & 0.6743 & 100.9 \\ \hline
ResNet-50   & 3 & 0.9088 & 0.8922 & 0.8197 & 0.6545 & 101.5 \\ \hline
ResNet-50   & 4 & 0.9012 & 0.8698 & 0.7923 & 0.5926 &  98.9 \\ \hline
ResNet-50   & 5 & 0.9052 & 0.8942 & 0.8251 & 0.6570 & 101.7 \\ \hline
MobileNetV3-Large & 2 & 0.8949 & 0.8869 & 0.8361 & 0.6468 &  97.6 \\ \hline
MobileNetV3-Large & 3 & 0.9040 & 0.8672 & 0.8169 & 0.6245 &  95.5 \\ \hline
MobileNetV3-Large & 4 & 0.9140 & 0.9017 & 0.8361 & 0.6450 &  99.3 \\ \hline
MobileNetV3-Large & 5 & 0.9024 & 0.8837 & 0.8197 & 0.6118 &  97.3 \\ \hline
Swin-T     & 2 & 0.8964 & 0.8782 & 0.8060 & 0.6302 &  97.3 \\ \hline
Swin-T     & 3 & 0.9097 & 0.8924 & 0.8224 & 0.6392 &  98.9 \\ \hline
Swin-T     & 4 & 0.9250 & 0.9093 & 0.8443 & 0.6851 & 100.8 \\ \hline
Swin-T     & 5 & 0.9126 & 0.8926 & 0.8142 & 0.6281 &  98.9 \\ \hline
\end{tabular}
\label{tab:alpha_full}
\end{table}

Values above 100\% indicate that the student's observed test QWK exceeded the corresponding teacher's observed test QWK; they should not be interpreted as literal recovery beyond the teacher's generalisation ability.

\subsection{Multi-seed robustness verification}

Six of the twenty-four (teacher, $T$, $\alpha$) configurations were repeated at three seeds (42, 123, and 999). The repeated configurations were selected after the single-seed screening using two practical criteria: (i) the three smallest observed alpha margins in Table~\ref{tab:alpha_grid}, corresponding to ResNet-50 at $T{=}3$, ResNet-50 at $T{=}4$, and Swin-T at $T{=}4$; and (ii) the low seed-42 test QWK accompanied by a large validation--test gap at ResNet-50 $T{=}4$. These criteria led to repeated evaluation of ResNet-50 at $T{=}3$ and $T{=}4$ and Swin-T at $T{=}4$, each at both alpha values. The remaining eighteen configurations were evaluated only at seed 42.

Individual repeated-seed results are given in Tables~\ref{tab:seeds_swint} and~\ref{tab:seeds_resnet}, and configuration means are given in Table~\ref{tab:seed_means}, with filtered means reported alongside as a sensitivity analysis. The seed-42 validation--test gap at ResNet-50 $T{=}4$ was approximately seven times the mean gap of the two counterpart seeds at the same configuration: 0.0321 versus 0.0047 for $\alpha{=}0.3$, and 0.0314 versus 0.0047 for $\alpha{=}0.7$.

\begin{table}[!htbp]
\centering
\caption{\textbf{Repeated-seed results, Swin-T$\to$MobileNetV3-Small, $T{=}4$.}}
\begin{tabular}{|c|c|r|r|r|r|}
\hline
\textbf{$\alpha$} & \textbf{Seed} & \textbf{Val QWK} & \textbf{Test QWK} & \textbf{Acc.} & \textbf{macro-F1} \\ \hline
0.3 & 42  & 0.9050 & 0.9065 & 0.8251 & 0.6620 \\ \hline
0.3 & 123 & 0.9030 & 0.8966 & 0.7978 & 0.6341 \\ \hline
0.3 & 999 & 0.9198 & 0.8984 & 0.8169 & 0.6383 \\ \hline
0.7 & 42  & 0.9250 & 0.9093 & 0.8443 & 0.6851 \\ \hline
0.7 & 123 & 0.9098 & 0.8999 & 0.8224 & 0.6410 \\ \hline
0.7 & 999 & 0.9173 & 0.8979 & 0.8361 & 0.6580 \\ \hline
\end{tabular}
\label{tab:seeds_swint}
\end{table}

\begin{table}[!htbp]
\centering
\caption{\textbf{Repeated-seed results, ResNet-50$\to$MobileNetV3-Small, $T{=}3$ and $T{=}4$.} The dagger marks runs whose validation--test QWK gap is approximately seven times the mean gap of their two counterpart seeds at the same configuration; these runs are retained in the reported means.}
\begin{tabular}{|c|c|c|r|r|r|r|}
\hline
\textbf{$T$} & \textbf{$\alpha$} & \textbf{Seed} & \textbf{Val QWK} & \textbf{Test QWK} & \textbf{Val$-$Test} & \textbf{macro-F1} \\ \hline
3 & 0.3 & 42  & 0.9048 & 0.8939 & 0.0109 & 0.6425 \\ \hline
3 & 0.3 & 123 & 0.9082 & 0.9075 & 0.0007 & 0.6681 \\ \hline
3 & 0.3 & 999 & 0.8996 & 0.8856 & 0.0140 & 0.6327 \\ \hline
3 & 0.7 & 42  & 0.9088 & 0.8922 & 0.0166 & 0.6545 \\ \hline
3 & 0.7 & 123 & 0.9139 & 0.8915 & 0.0224 & 0.6729 \\ \hline
3 & 0.7 & 999 & 0.9136 & 0.8813 & 0.0323 & 0.6551 \\ \hline
4 & 0.3 & 42$^\dagger$ & 0.8987 & 0.8666 & 0.0321 & 0.5965 \\ \hline
4 & 0.3 & 123 & 0.9053 & 0.9003 & 0.0050 & 0.6677 \\ \hline
4 & 0.3 & 999 & 0.9078 & 0.9034 & 0.0044 & 0.6482 \\ \hline
4 & 0.7 & 42$^\dagger$ & 0.9012 & 0.8698 & 0.0314 & 0.5926 \\ \hline
4 & 0.7 & 123 & 0.9146 & 0.9081 & 0.0065 & 0.6739 \\ \hline
4 & 0.7 & 999 & 0.9078 & 0.9049 & 0.0029 & 0.6399 \\ \hline
\end{tabular}
\label{tab:seeds_resnet}
\end{table}

\begin{table}[!htbp]
\centering
\caption{\textbf{Mean test QWK over the repeated configurations.} Primary means include all three seeds; the final column repeats the calculation with the flagged seed-42 runs at ResNet-50 $T{=}4$ excluded, as a sensitivity analysis.}
\begin{tabular}{|l|c|c|r|r|}
\hline
\textbf{Teacher} & \textbf{$T$} & \textbf{$\alpha$} & \textbf{Mean (3 seeds)} & \textbf{Mean (excl.\ flagged)} \\ \hline
ResNet-50 & 3 & 0.3 & 0.8957 & -- \\ \hline
ResNet-50 & 3 & 0.7 & 0.8883 & -- \\ \hline
ResNet-50 & 4 & 0.3 & 0.8901 & 0.9019 \\ \hline
ResNet-50 & 4 & 0.7 & 0.8943 & 0.9065 \\ \hline
Swin-T    & 4 & 0.3 & 0.9005 & -- \\ \hline
Swin-T    & 4 & 0.7 & 0.9024 & -- \\ \hline
\multicolumn{3}{|l|}{No-KD baseline, 3-seed mean} & 0.8804 & -- \\ \hline
\end{tabular}
\label{tab:seed_means}
\end{table}

\begin{figure}[!htbp]
\centering
\includegraphics[width=0.9\linewidth]{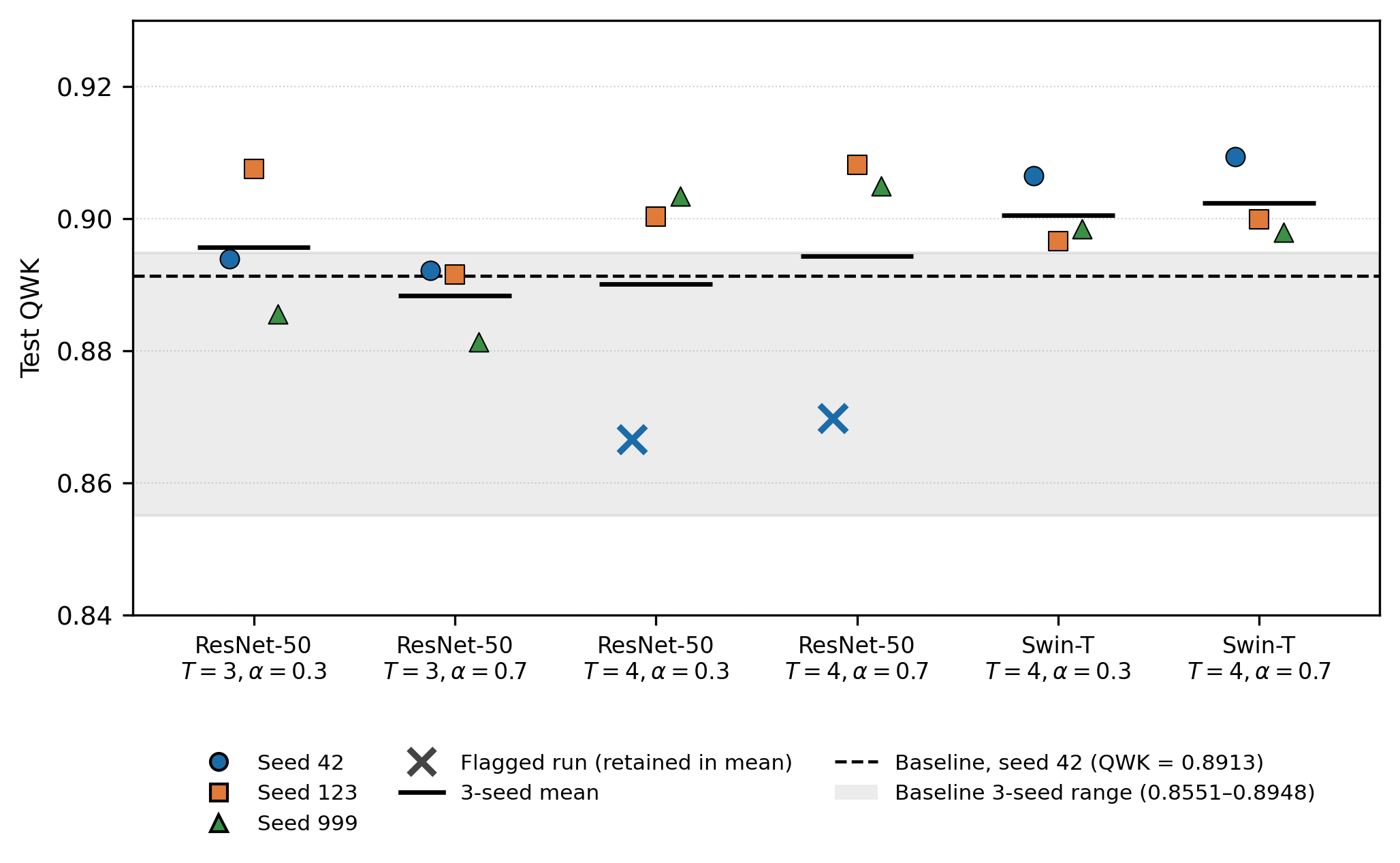}
\caption{\textbf{Repeated-seed test QWK for the six verified configurations.}
Individual test QWK values at seeds 42, 123 and 999 for ResNet-50$\to$MobileNetV3-Small ($T{=}3$ and $T{=}4$) and Swin-T$\to$MobileNetV3-Small ($T{=}4$), each at $\alpha{=}0.3$ and $\alpha{=}0.7$ (values as in Tables~\ref{tab:seeds_swint} and~\ref{tab:seeds_resnet}). Horizontal bars show the three-seed mean. Crosses mark the two flagged seed-42 runs at ResNet-50 $T{=}4$, which are retained in the means. The dashed line is the seed-42 no-KD baseline (QWK 0.8913) and the shaded band is the three-seed baseline range (0.8551--0.8948).}
\label{fig:seed_verification}
\end{figure}

Across the repeated runs, the Swin-T $T{=}4$ configuration produced test QWK values from 0.8966 to 0.9093, exceeding the highest observed no-KD baseline value of 0.8948 in all six runs. The repeated-seed means showed only small alpha differences: $\alpha{=}0.7$ was higher for Swin-T at $T{=}4$ and ResNet-50 at $T{=}4$, whereas $\alpha{=}0.3$ was higher for ResNet-50 at $T{=}3$. These margins were small relative to the observed baseline seed range and do not support a general alpha preference. The low seed-42 result for ResNet-50 at $T{=}4$ was not reproduced at the two additional seeds, suggesting seed-dependent variation; the remaining single-seed shortfalls were not repeated. Fig~\ref{fig:seed_verification} shows the individual seed values underlying Tables~\ref{tab:seeds_swint} and~\ref{tab:seeds_resnet}, together with the flagged runs. When the flagged runs are excluded, the ResNet-50 $T{=}4$ means (0.9019 and 0.9065) are comparable to the Swin-T $T{=}4$ means (0.9005 and 0.9024), so the repeated-seed data do not separate these two teachers.

\subsection{Best-configuration summary}

Table~\ref{tab:best_config} reports, for each teacher, the configuration with the highest single-seed test QWK identified in the temperature and alpha analyses above, together with its seed-verification status.

\begin{table}[!htbp]
\centering
\caption{\textbf{Highest observed screening result per teacher.} Comp.: teacher-to-student parameter ratio. Ratios are computed from unrounded parameter counts. Test QWK values are single-seed screening maxima over the $(T,\alpha)$ grid.}
\begin{tabular}{|l|c|c|r|r|l|}
\hline
\textbf{Teacher} & \textbf{$T$} & \textbf{$\alpha$} & \textbf{Test QWK} & \textbf{Comp.} & \textbf{Verification status} \\ \hline
ResNet-50   & 5 & 0.3 & 0.9090 & 15.44$\times$ & single seed only \\ \hline
MobileNetV3-Large & 2 & 0.3 & 0.9066 &  2.76$\times$ & single seed only \\ \hline
Swin-T      & 4 & 0.7 & 0.9093 & 18.07$\times$ & 3 seeds; mean 0.9024 (Table~\ref{tab:seed_means}) \\ \hline
\end{tabular}
\label{tab:best_config}
\end{table}

\subsection{Per-class analysis}

Table~\ref{tab:perclass} reports per-class F1 for the baseline and the best single-seed student identified per pipeline in Table~\ref{tab:best_config}.

\begin{table}[!htbp]
\footnotesize
\centering
\caption{\textbf{Per-class F1 on the test set: seed-42 no-KD baseline against the highest observed student per teacher pipeline (Table~\ref{tab:best_config}).} All rows are single-seed results; the configurations were identified by their test QWK, so these values describe the selected runs rather than providing independent per-class estimates. MNV3-S, MobileNetV3-Small; MNV3-L, MobileNetV3-Large.}
\begin{adjustwidth}{-0.5in}{-0.5in}
\centering
\begin{tabular}{|l|r|r|r|r|r|r|r|}
\hline
\textbf{Model} & \textbf{F1$_0$} & \textbf{F1$_1$} & \textbf{F1$_2$} & \textbf{F1$_3$} & \textbf{F1$_4$} & \textbf{macro-F1} & \textbf{QWK} \\ \hline
Baseline (No KD) & 0.9823 & 0.5000 & 0.7407 & 0.3333 & 0.5385 & 0.6190 & 0.8913 \\ \hline
ResNet-50$\to$MNV3-S ($T{=}5$) & 0.9796 & 0.5479 & 0.7066 & 0.3902 & \textbf{0.6441} & 0.6537 & 0.9090 \\ \hline
MNV3-L$\to$MNV3-S ($T{=}2$) & \textbf{0.9874} & 0.5970 & \textbf{0.7845} & 0.3784 & 0.6000 & 0.6695 & 0.9066 \\ \hline
Swin-T$\to$MNV3-S ($T{=}4,\alpha{=}0.7$) & 0.9823 & \textbf{0.6269} & 0.7717 & \textbf{0.4324} & 0.6122 & \textbf{0.6851} & \textbf{0.9093} \\ \hline
\end{tabular}
\end{adjustwidth}
\label{tab:perclass}
\end{table}

\begin{figure}[!htbp]
\centering
\includegraphics[width=0.95\linewidth]{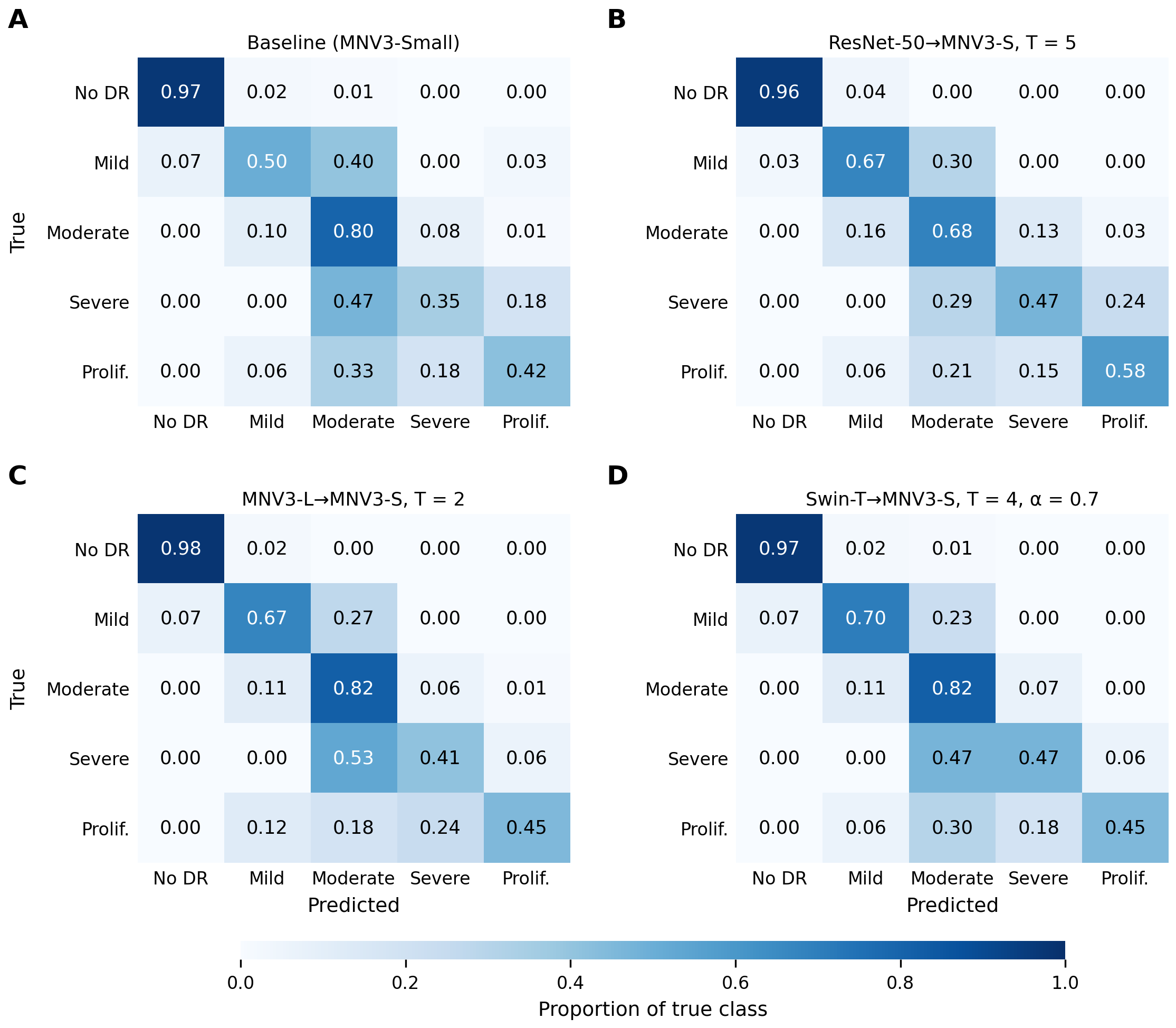}
\caption{\textbf{Row-normalised confusion matrices on the test set.}
(A) No-KD MobileNetV3-Small baseline (seed 42). (B) ResNet-50$\to$MobileNetV3-Small, $T{=}5$, $\alpha{=}0.3$. (C) MobileNetV3-Large$\to$MobileNetV3-Small, $T{=}2$, $\alpha{=}0.3$. (D) Swin-T$\to$MobileNetV3-Small, $T{=}4$, $\alpha{=}0.7$. Each row sums to 1 over the true class; the test set contains 199, 30, 87, 17 and 33 images of No DR, Mild, Moderate, Severe and Proliferative DR, respectively. Panels B--D show the highest-QWK student per teacher, selected on test QWK (Table~\ref{tab:best_config}).}
\label{fig:confmat}
\end{figure}

Fig~\ref{fig:confmat} shows normalised confusion matrices for the baseline and the three best-performing students.

\subsection{Efficiency}

Table~\ref{tab:efficiency} reports parameter count, model size and latency for the three teachers and the deployed student. Latency was measured at batch size 1 on GPU with synchronisation, as the mean of 100 timed forward passes after warm-up. At this batch size, per-call overhead contributes a substantial share of the measured time and the architectures are not well separated: the 23.52~M-parameter ResNet-50 (5.61~ms) is faster than the 4.21~M-parameter MobileNetV3-Large (6.33~ms). The efficiency argument for the distilled student therefore rests on its 1.52~M parameters and 5.94~MB footprint, which are 18.07$\times$ and 17.72$\times$ smaller than the Swin-T teacher respectively, rather than on the latency column. On-device CPU benchmarking was not performed.

\begin{table}[!htbp]
\centering
\caption{\textbf{Model efficiency.} Latency: mean of 100 single-image forward passes, GPU, batch size 1.}
\begin{tabular}{|l|r|r|r|l|}
\hline
\textbf{Model} & \textbf{Params (M)} & \textbf{Size (MB)} & \textbf{Lat.\ (ms)} & \textbf{Role} \\ \hline
ResNet-50   & 23.52 &  90.02 &  5.61 & Teacher \\ \hline
MobileNetV3-Large &  4.21 &  16.25 &  6.33 & Teacher \\ \hline
Swin-T     & 27.52 & 105.28 & 12.22 & Teacher \\ \hline
MobileNetV3-Small &  1.52 &   5.94 &  5.00 & Student (deployed) \\ \hline
\end{tabular}
\label{tab:efficiency}
\end{table}

\section{Discussion}

This section addresses the three questions posed in the Introduction in turn: the influence of teacher architecture, the sensitivity of distillation to $T$ and $\alpha$, and the per-class behaviour of the distilled students.

\subsection{Distillation into a compact student was associated with improved ordinal agreement}
The most consistently supported result in this study is the comparison between the repeated Swin-T $T{=}4$ students and the no-KD baseline. All six repeated Swin-T runs, spanning three seeds at each of two $\alpha$ values, produced a test QWK above the highest of the three baseline seeds. The three-seed mean at $\alpha{=}0.7$ (0.9024) exceeded the three-seed baseline mean (0.8804) by 0.0220 QWK. Because this ordering held at every reported seed for both configurations, it did not depend on the choice of baseline reference run. Across the wider single-seed grid, ten of twelve configurations at $\alpha{=}0.3$ exceeded the three-seed baseline mean of 0.8804, although this is a descriptive comparison because those grid cells were not repeated. The distilled student reaching this level uses 1.52~M parameters compared with 27.52~M for the Swin-T teacher, an 18.07$\times$ reduction. Its three-seed mean QWK of 0.9024 matches the reported QWK of the single Swin-T teacher checkpoint to four decimal places. The Swin-T $T{=}4$, $\alpha{=}0.7$ configuration also had the highest validation QWK in the whole grid (0.9250) and the highest three-seed mean validation QWK among the six repeated configurations (0.9174), so it would have been selected without reference to test data.

\subsection{Teacher architecture influenced the error profile more than the aggregate score}
The three teachers differed in architecture and parameter count, yet their highest observed single-seed student QWK values differed by only 0.0027 (0.9066--0.9093; Table~\ref{tab:best_config}). Because these values were obtained through test-set screening, they should not be treated as independent estimates of generalisation. The per-class results suggest that the differences lie elsewhere. The Swin-T student recorded the highest macro-F1 (0.6851), while the ResNet-50 student recorded the highest Proliferative DR F1 (0.6441), even though ResNet-50 was the lowest-QWK teacher. No pipeline led on every class. MobileNetV3-Large, the smallest teacher at 4.21~M parameters, recorded the highest teacher QWK (0.9084). Its pairing with MobileNetV3-Small represents only a 2.76$\times$ parameter reduction and is therefore better interpreted as a comparison between architecturally similar models than as the strongest compression setting. The limited size and class imbalance of the APTOS 2019 training split may limit the extent to which higher-capacity backbones realise a capacity advantage; this explanation was not tested directly.

\subsection{Temperature and hard-label weighting interact, and neither admits a single default}
At $\alpha{=}0.3$, the temperature associated with each teacher's highest observed test QWK differed across architectures ($T{=}2$, $T{=}4$, and $T{=}5$ for MobileNetV3-Large, Swin-T, and ResNet-50, respectively), with a within-teacher spread across temperatures of 0.0244--0.0424 QWK. This ordering did not carry over to $\alpha{=}0.7$, where MobileNetV3-Large's highest observed value also fell at $T{=}4$. The $\alpha$ preference itself tracked temperature: at $T{=}2$ and $T{=}5$ every teacher favoured $\alpha{=}0.3$, whereas at $T{=}4$ every teacher favoured $\alpha{=}0.7$; $T{=}3$ was the sole point of disagreement. Together, these results suggest that temperature and $\alpha$ interact under the tested configurations, so fixed values cannot be assumed to transfer consistently across teacher architectures. A possible mechanism is that the soft-target term in Eq~(\ref{eq:kd_loss}) is scaled by $T^2$, causing the effective balance of gradient contributions to vary with temperature even when $\alpha$ is held fixed. Disentangling this scaling from any genuine interaction would require an experiment not performed here.

\subsection{Repeated-seed results did not establish a universal \texorpdfstring{$\alpha$}{alpha} preference}
The three-seed means favoured $\alpha{=}0.7$ for Swin-T at $T{=}4$ and ResNet-50 at $T{=}4$, whereas they favoured $\alpha{=}0.3$ for ResNet-50 at $T{=}3$. These mean differences were small relative to the observed baseline seed range of 0.0397, indicating configuration-dependent tendencies rather than a general preference for either $\alpha$ value. The low seed-42 result for ResNet-50 at $T{=}4$ was not reproduced at the two additional seeds, which is consistent with seed-dependent variation, although three seeds are insufficient for a firm attribution. The remaining single-seed shortfalls were not repeated and remain unverified. For the Swin-T $T{=}4$, $\alpha{=}0.7$ configuration, the highest single-seed value (0.9093) was 0.0069 QWK above its three-seed mean (0.9024), illustrating how a single observed maximum can differ from a repeated-seed summary.

\subsection{Per-class behaviour differed mainly across the DR grades}
All students and teachers classified No DR at F1 above 0.97, so the largest differences between configurations occurred in the disease grades. Mild DR showed the largest observed improvement: the Swin-T student reached F1 0.6269 compared with the baseline range of 0.4400--0.5357 across seeds, exceeding the highest baseline-seed value by 0.0912. Because per-class scores were not recorded for the repeated student seeds, this comparison remains a single-seed observation and should be interpreted cautiously. Among the reported teacher and student configurations, Severe DR had the lowest per-class F1, with a maximum of 0.4324. With only 30 Mild and 17 Severe DR images in the test set, these per-class estimates are sensitive to a small number of individual predictions.

\subsection{Comparison with published work}
Table~\ref{tab:sota} places these results alongside recent APTOS 2019 studies, with protocol differences stated explicitly, because such comparisons are informative only where task, split, and metric match. The most directly comparable listed study reports five-class ordinal QWK of $0.923 \pm 0.008$ under five-fold cross-validation and 0.919 on a held-out split, both above the value obtained here. That study also reports a variance estimate, whereas the present single-split design does not provide a variance estimate for the grid as a whole. A further study reports 99.04\% five-class accuracy without a stated split or variance estimate; accuracy on an unspecified split and QWK on a held-out split are not directly comparable, and the row is included for completeness rather than as a benchmark. We make no claim of state-of-the-art performance. The contribution of this work lies in the systematic comparison of CNN and transformer teachers distilled into a common student across a temperature--weighting grid under a single protocol, together with seed-level characterisation of the variability of selected comparisons.

\begin{table}[!htbp]
\footnotesize
\centering
\caption{\textbf{Comparison with selected recent studies using the APTOS 2019 dataset.} Direct comparison is valid only for studies using the same classification task, dataset split, and evaluation metric. Protocol differences are reported explicitly.}
\begin{tabular}{|p{3.3cm}|p{2.6cm}|p{2.0cm}|p{4.4cm}|p{1.9cm}|}
\hline
\textbf{Method} & \textbf{Venue (Year)} & \textbf{Task} & \textbf{Metric / Protocol} & \textbf{Value} \\ \hline
Islam et al.~\cite{app132212397} & Appl. Sci. (2023) & Five-class & Accuracy (split unspecified) & 99.04\% \\ \hline
Dharrao et al.~\cite{dharrao2025efficientnet} & Discov. Appl. Sci. (2025) & Five-class ordinal & QWK, five-fold CV & $0.923 \pm 0.008$ \\ \hline
Dharrao et al.~\cite{dharrao2025efficientnet} & Discov. Appl. Sci. (2025) & Five-class ordinal & QWK, held-out test set & 0.919 \\ \hline
\textbf{This work: Swin-T $\rightarrow$ MobileNetV3-Small KD} & -- & Five-class ordinal & QWK, single held-out split; highest observed single-seed result over a 24-cell screening grid, with the three-seed mean for the same configuration in parentheses & \textbf{0.9093} (0.9024) \\ \hline
\end{tabular}
\label{tab:sota}
\end{table}

\subsection{Scope and future work}

This study used a single dataset (APTOS 2019), a single third-party train/validation/test split whose construction is undocumented, and one training protocol. The findings therefore do not establish generalisation beyond this split, and no claim of clinical readiness is made.

Most grid configurations were evaluated at one seed, whereas only selected configurations received repeated-seed verification. The observed baseline seed variation therefore warrants cautious interpretation of small single-seed differences. In addition, each teacher was represented by one trained checkpoint, and the imbalanced test set limits the precision of estimates for the rarer DR grades.

Future work should evaluate the selected configurations across more seeds and independent DR datasets, examine deployment performance on CPU or edge hardware, and compare KD with complementary compression methods such as pruning and quantisation.

\section{Conclusion}

This study evaluated offline knowledge distillation for five-class diabetic retinopathy grading on the APTOS~2019 dataset using ResNet-50, MobileNetV3-Large, and Swin-T teachers and a MobileNetV3-Small student. Distillation produced competitive student performance under the tested protocol, with the highest observed single-seed QWK reaching 0.9093 for the Swin-T $\rightarrow$ MobileNetV3-Small configuration at $T{=}4$ and $\alpha{=}0.7$. Across three seeds, the same configuration achieved a mean QWK of 0.9024, while using 1.52~M parameters and a 5.94~MB model footprint. The temperature--alpha analysis indicated that their effects depended on the teacher and configuration rather than supporting a universal setting. The results therefore support knowledge distillation as a practical model-compression strategy under this experimental protocol, but not as evidence of universal superiority or clinical readiness. The observed seed variability, single-checkpoint teacher comparisons, single-split evaluation, and absence of external validation motivate further evaluation across additional seeds, independent datasets, and deployment-relevant hardware before clinical use is considered.

\section*{Acknowledgments}

The authors used AI-assisted language tools (GPT 5.6 and Anthropic Claude Opus 5) for grammar, readability, and language editing during manuscript preparation. These tools were not used to generate data, conduct experiments, perform analyses, or make scientific conclusions. All content was reviewed and approved by the authors, who take responsibility for the final manuscript.

\bibliographystyle{unsrt}
\bibliography{references}

@article{yau2012global,
  title={Global prevalence and major risk factors of diabetic retinopathy},
  author={Yau, Joanne WY and Rogers, Sophie L and Kawasaki, Ryo and Lamoureux, Ecosse L and Kowalski, Jonathan W and Bek, Toke and Chen, Shih-Jen and Dekker, Jacqueline M and Fletcher, Astrid and Grauslund, Jakob and others},
  journal={Diabetes Care},
  volume={35},
  number={3},
  pages={556--564},
  year={2012},
  publisher={Am Diabetes Assoc}
}

@article{gulshan2016development,
  title   = {Development and validation of a deep learning algorithm for detection of diabetic retinopathy in retinal fundus photographs},
  author  = {Gulshan, Varun and Peng, Lily and Coram, Marc and Stumpe, Martin C. and Wu, Derek and Narayanaswamy, Arunachalam and Venugopalan, Subhashini and Widner, Kasumi and Madams, Tom and Cuadros, Jorge and Kim, Ramasamy and Raman, Rajiv and Nelson, Philip C. and Mega, Jessica L. and Webster, Dale R.},
  journal = {JAMA},
  volume  = {316},
  number  = {22},
  pages   = {2402--2410},
  year    = {2016},
  doi     = {10.1001/jama.2016.17216}
}

@article{lam2018automated,
  title   = {Automated Detection of Diabetic Retinopathy Using Deep Learning},
  author  = {Lam, Carson and Yi, Darvin and Guo, Margaret and Lindsey, Tony},
  journal = {AMIA Jt Summits Transl Sci Proc},
  volume  = {2017},
  pages   = {147--155},
  year    = {2018},
  pmid    = {29888061},
  pmcid   = {PMC5961805}
}

@inproceedings{joshi2025mobile,
  title={Mobile-Ready Automated Triage of Diabetic Retinopathy Using Digital Fundus Images},
  author={Joshi, Aadi and Sharma, Manav S and Rathod, Vijay Uttam and Sawant, Ashlesha and Musale, Prajakta and Kalamkar, Asmita B},
  booktitle={International Conference on Computational Intelligence},
  pages={190--198},
  year={2025},
  organization={Springer}
}

@misc{hinton2015distilling,
  title         = {Distilling the Knowledge in a Neural Network},
  author        = {Geoffrey Hinton and Oriol Vinyals and Jeff Dean},
  year          = {2015},
  eprint        = {1503.02531},
  archivePrefix = {arXiv},
  primaryClass  = {stat.ML},
  howpublished  = {arXiv preprint arXiv:1503.02531}
}

@inproceedings{beyer2022goodteacher,
  author={Beyer, Lucas and Zhai, Xiaohua and Royer, Amélie and Markeeva, Larisa and Anil, Rohan and Kolesnikov, Alexander},
  booktitle={2022 IEEE/CVF Conference on Computer Vision and Pattern Recognition (CVPR)}, 
  title={Knowledge distillation: A good teacher is patient and consistent}, 
  year={2022},
  volume={},
  number={},
  pages={10915-10924},
  doi={10.1109/CVPR52688.2022.01065}
}

@article{abbasi2021classification,
  author  = {Abbasi, Sajjad and Hajabdollahi, Mohsen and Khadivi, Pejman and
             Karimi, Nader and Roshandel, Roshanak and Shirani, Shahram and
             Samavi, Shadrokh},
  title   = {Classification of Diabetic Retinopathy Using Unlabeled Data and
             Knowledge Distillation},
  journal = {Artif Intell Med},
  year    = {2021},
  volume  = {121},
  pages   = {102176},
  doi     = {10.1016/j.artmed.2021.102176}
}

@article{math12162543,
  title   = {Computer-Aided Diagnosis of Diabetic Retinopathy Lesions
             Based on Knowledge Distillation in Fundus Images},
  author  = {Moya-Albor, Ernesto and Lopez-Figueroa, Alberto and
             Jacome-Herrera, Sebastian and Renza, Diego and Brieva, Jorge},
  journal = {Mathematics},
  volume  = {12},
  number  = {16},
  pages   = {2543},
  year    = {2024},
  doi     = {10.3390/math12162543},
  publisher = {MDPI}
}

@article{app132212397,
  title   = {Toward Lightweight Diabetic Retinopathy Classification:
             {A} Knowledge Distillation Approach for Resource-Constrained Settings},
  author  = {Islam, Niful and Jony, Md. Mehedi Hasan and Hasan, Emam and
             Sutradhar, Sunny and Rahman, Atikur and Islam, Md. Motaharul},
  journal = {Appl Sci},
  volume  = {13},
  number  = {22},
  pages   = {12397},
  year    = {2023},
  doi     = {10.3390/app132212397},
  publisher = {MDPI}
}

@inproceedings{shafi2024ijcnn,
  title     = {Enhancing Diabetic Retinopathy Detection Through Transformer
               Based Knowledge Distillation and Explainable {AI}},
  author    = {Shafi, A. A. and Hossain Shawon, M. and Shahid, N. and
               Rahman, R. and Khan, R.},
  booktitle = {2024 International Joint Conference on Neural Networks (IJCNN)},
  address   = {Yokohama, Japan},
  pages     = {1--8},
  year      = {2024},
  doi       = {10.1109/IJCNN60899.2024.10650642}
}

@article{wang2024lesion,
  author  = {Wang, Yaqi and Hou, Qingshan and Cao, Peng and
             Yang, Jinzhu and Za{\"i}ane, Osmar R.},
  title   = {Lesion-Aware Knowledge Distillation for Diabetic Retinopathy
             Lesion Segmentation},
  journal = {Appl Intell},
  year    = {2024},
  volume  = {54},
  number  = {2},
  pages   = {1937--1956},
  doi     = {10.1007/s10489-024-05274-8}
}

@article{roychowdhury2014dream,
  author  = {Roychowdhury, Sohini and Koozekanani, Dara D. and Parhi, Keshab K.},
  title   = {{DREAM}: Diabetic Retinopathy Analysis Using Machine Learning},
  journal = {IEEE J Biomed Health Inform},
  year    = {2014},
  volume  = {18},
  number  = {5},
  pages   = {1717--1728},
  month   = sep,
  doi     = {10.1109/JBHI.2013.2294635},
  pmid    = {25192577}
}

@misc{haloi2015improved,
  author        = {Haloi, Mrinal},
  title         = {Improved Microaneurysm Detection Using Deep Neural Networks},
  year          = {2015},
  eprint        = {1505.04424},
  archivePrefix = {arXiv},
  primaryClass  = {cs.CV},
  howpublished = {arXiv preprint arXiv:1505.04424}
}

@misc{kaggle2015,
  author       = {{Kaggle}},
  title        = {Diabetic Retinopathy Detection},
  year         = {2015},
  howpublished = {\url{https://www.kaggle.com/competitions/diabetic-retinopathy-detection}},
  note         = {Kaggle competition}
}

@article{pratt2016convolutional,
  author  = {Pratt, Harry and Coenen, Frans and Broadbent, Deborah M. and
             Harding, Simon P. and Zheng, Yalin},
  title   = {Convolutional Neural Networks for Diabetic Retinopathy},
  journal = {Procedia Comput Sci},
  year    = {2016},
  volume  = {90},
  pages   = {200--205},
  doi     = {10.1016/j.procs.2016.07.090}
}

@article{abramoff2016improved,
  author  = {Abr{\`a}moff, Michael David and Lou, Yiyue and Erginay, Ali and
             Clarida, Warren and Amelon, Ryan and Folk, James C. and
             Niemeijer, Meindert},
  title   = {Improved Automated Detection of Diabetic Retinopathy on a Publicly
             Available Dataset Through Integration of Deep Learning},
  journal = {Invest Ophthalmol Vis Sci},
  year    = {2016},
  volume  = {57},
  number  = {13},
  pages   = {5200--5206},
  doi     = {10.1167/iovs.16-19964}
}

@article{takahashi2017applying,
  author  = {Takahashi, Hidenori and Tampo, Hironobu and Arai, Yusuke and
             Inoue, Yuji and Kawashima, Hidetoshi},
  title   = {Applying Artificial Intelligence to Disease Staging: Deep Learning
             for Improved Staging of Diabetic Retinopathy},
  journal = {PLoS ONE},
  year    = {2017},
  volume  = {12},
  number  = {6},
  pages   = {e0179790},
  doi     = {10.1371/journal.pone.0179790}
}

@article{ting2017development,
  author  = {Ting, Daniel Shu Wei and Cheung, Carol Yim-Lui and Lim, Gilbert and
             Tan, Gavin Siew Wei and Quang, Nguyen D. and Gan, Alfred and others},
  title   = {Development and Validation of a Deep Learning System for Diabetic
             Retinopathy and Related Eye Diseases Using Retinal Images from
             Multiethnic Populations with Diabetes},
  journal = {JAMA},
  year    = {2017},
  volume  = {318},
  number  = {22},
  pages   = {2211--2223},
  doi     = {10.1001/jama.2017.18152}
}

@article{orlando2018ensemble,
  author  = {Orlando, Jos{\'e} Ignacio and Prokofyeva, Elena and
             Del Fresno, Mariana and Blaschko, Matthew B.},
  title   = {An Ensemble Deep Learning Based Approach for Red Lesion Detection
             in Fundus Images},
  journal = {Comput Methods Programs Biomed},
  year    = {2018},
  volume  = {153},
  pages   = {115--127},
  doi     = {10.1016/j.cmpb.2017.11.014}
}

@inproceedings{wang2018diabetic,
  author    = {Wang, Zhiguang and Yang, Jianbo},
  title     = {Diabetic Retinopathy Detection via Deep Convolutional Networks for
               Discriminative Localization and Visual Explanation},
  booktitle = {AAAI Workshops},
  year      = {2018},
  pages     = {514--521}
}

@misc{aptos2019,
  author       = {{Asia Pacific Tele-Ophthalmology Society (APTOS)}},
  title        = {{APTOS 2019 Blindness Detection}},
  year         = {2019},
  howpublished = {Kaggle competition dataset},
  url          = {https://www.kaggle.com/c/aptos2019-blindness-detection}
}

@misc{romero2014fitnets,
  author        = {Romero, Adriana and Ballas, Nicolas and Kahou, Samira Ebrahimi and                           Chassang, Antoine and Gatta, Carlo and Bengio, Yoshua},
  title         = {{FitNets}: Hints for Thin Deep Nets},
  year          = {2014},
  eprint        = {1412.6550},
  archivePrefix = {arXiv},
  primaryClass  = {cs.LG},
  howpublished = {arXiv preprint arXiv:1412.6550}
}

@misc{zagoruyko2016paying,
  author        = {Zagoruyko, Sergey and Komodakis, Nikos},
  title         = {Paying More Attention to Attention: Improving the Performance of
                   Convolutional Neural Networks via Attention Transfer},
  year          = {2016},
  eprint        = {1612.03928},
  archivePrefix = {arXiv},
  primaryClass  = {cs.CV},
  howpublished = {arXiv preprint arXiv:1612.03928}
}

@inproceedings{park2019relational,
  author    = {Park, Wonpyo and Kim, Dongju and Lu, Yan and Cho, Minsu},
  title     = {Relational Knowledge Distillation},
  booktitle = {2019 IEEE/CVF Conference on Computer Vision and Pattern Recognition (CVPR)},
  year      = {2019},
  pages     = {3962--3971},
  doi       = {10.1109/CVPR.2019.00409}
}

@misc{tian2019contrastive,
  author        = {Tian, Yonglong and Krishnan, Dilip and Isola, Phillip},
  title         = {Contrastive Representation Distillation},
  year          = {2019},
  eprint        = {1910.10699},
  archivePrefix = {arXiv},
  primaryClass  = {cs.LG},
  howpublished = {arXiv preprint arXiv:1910.10699}
}

@inproceedings{touvron2021training,
  author    = {Touvron, Hugo and Cord, Matthieu and Douze, Matthijs and
               Massa, Francisco and Sablayrolles, Alexandre and J{\'e}gou, Herv{\'e}},
  title     = {Training Data-Efficient Image Transformers \& Distillation
               Through Attention},
  booktitle = {International Conference on Machine Learning},
  year      = {2021},
  pages     = {10347--10357}
}

@article{wilkinson2003proposed,
  title={Proposed international clinical diabetic retinopathy and diabetic macular edema disease severity scales},
  author={Wilkinson, Charles P and Ferris III, Frederick L and Klein, Ronald E and Lee, Paul P and Agardh, Carl David and Davis, Matthew and Dills, Diana and Kampik, Anselm and Pararajasegaram, Rangasamy and Verdaguer, Juan T and others},
  journal={Ophthalmology},
  volume={110},
  number={9},
  pages={1677--1682},
  year={2003},
  publisher={Elsevier}
}

@article{rohan1989prevention,
  title   = {Prevention of Blindness by Screening for Diabetic Retinopathy:
             A Quantitative Assessment},
  author  = {Rohan, Thomas E. and Frost, Chris and Wald, Nicholas J.},
  journal = {BMJ},
  year    = {1989},
  volume  = {299},
  number  = {6709},
  pages   = {1198--1201},
  doi     = {10.1136/bmj.299.6709.1198} 
}

@inproceedings{howard2019searching,
  title={Searching for {MobileNetV3}},
  author={Howard, Andrew and Sandler, Mark and Chu, Grace and Chen, Liang-Chieh and Chen, Bo and Tan, Mingxing and Wang, Weijun and Zhu, Yukun and Pang, Ruoming and Vasudevan, Vijay and others},
  booktitle = {Proceedings of the IEEE/CVF International Conference on Computer Vision},
  pages={1314--1324},
  year={2019}
}

@article{zhang2025dual,
  author  = {Zhang, Menglin and Liu, Qi and Zhan, Jialei and
             Gao, Jinwen and Xie, Dong and Liu, Jialang},
  title   = {A Dual Attention and Multi-Scale Fusion Network for
             Diabetic Retinopathy Image Analysis},
  journal = {Front Med (Lausanne)},
  year    = {2025},
  volume  = {12},
  pages   = {1614046},
  doi     = {10.3389/fmed.2025.1614046},
  issn    = {2296-858X}
}

@article{dejene2025knowledge,
  author  = {Dejene, Fitsum Mesfin and Ayano, Yehualashet Megersa and
             Feyisa, Degaga Wolde and Debelee, Taye Girma and
             Mekonnen, Hiwot Taye and Gessesse, Girum Woldegebreal and
             Merga, Zelalem Chimdesa and Molla, Hasset Tamirat and
             Mulie, Destaw},
  title   = {Knowledge Distillation-Based Lightweight {MobileNet} Model
             for Diabetic Retinopathy Classification},
  journal = {Sci Rep},
  volume  = {16},
  number  = {1},
  pages   = {1181},
  year    = {2025},
  doi     = {10.1038/s41598-025-30893-7},
  pmid    = {41350373},
  pmcid   = {PMC12789564}
}

@article{dharrao2025efficientnet,
  author  = {Dharrao, Deepak and Dharrao, Madhuri and Patil, Shreyas and
             Salvin, Sangeeth and Ahire, Prashant and Dongre, Yashwant},
  title   = {{AI}-Driven Detection and Classification of Diabetic Retinopathy
             Stages Using {EfficientNetB0}},
  journal = {Discover Applied Sciences},
  year    = {2025},
  volume  = {7},
  pages   = {1400},
  doi     = {10.1007/s42452-025-07998-9}
}

@misc{aptos_split,
  author       = {{MariaHerreroT}},
  title        = {{APTOS-2019} dataset},
  howpublished = {Kaggle dataset, Version 3},
  year         = {2021},
  note         = {Available from: https://www.kaggle.com/datasets/mariaherrerot/aptos2019. Accessed 2026 Oct 19}
}

@inproceedings{he2016deep,
  title={Deep Residual Learning for Image Recognition},
  author={He, Kaiming and Zhang, Xiangyu and Ren, Shaoqing and Sun, Jian},
  booktitle={Proc. IEEE Conf. Comput. Vis. Pattern Recognit. (CVPR)},
  pages={770--778},
  year={2016},
  publisher={IEEE},
  doi={10.1109/CVPR.2016.90}
}

@inproceedings{liu2021swin,
  title={Swin transformer: Hierarchical vision transformer using shifted windows},
  author={Liu, Ze and Lin, Yutong and Cao, Yue and Hu, Han and Wei, Yixuan and Zhang, Zheng and Lin, Stephen and Guo, Baining},
  booktitle={2021 IEEE/CVF Int. Conf. Comput. Vis.(ICCV)},
  pages={9992--10002},
  year={2021},
  organization={IEEE},
  
  doi={10.1109/ICCV48922.2021.00986}
}

@article{loshchilov2017decoupled,
  title={Decoupled weight decay regularization},
  author={Loshchilov, Ilya and Hutter, Frank},
  journal={arXiv preprint arXiv:1711.05101},
  year={2017}
}

\end{document}